\documentclass[10pt,twocolumn,letterpaper]{article}

\usepackage[pagenumbers]{cvpr} % To force page numbers, e.g. for an arXiv version

\usepackage{multirow} 
\usepackage{booktabs}
\usepackage{graphicx}
\usepackage{pifont}
\usepackage[table]{xcolor}
\usepackage{colortbl}
\usepackage{booktabs}
\usepackage{multirow, hhline}
\usepackage{array}
\usepackage{arydshln} % 可选
\usepackage{booktabs} % optional but recommended
\usepackage{soul}       % 提供 \hl 命令
\usepackage{xcolor}     % 提供颜色支持

\newcommand{\customtitle}[2]{%
  \begin{center}
    {\large \textbf{#1}} \\[0.5em] % 第一行标题，粗体
    \vspace{1em} % 标题与正文之间的间距
  \end{center}
}

\definecolor{cvprblue}{rgb}{0.21,0.49,0.74}
\definecolor{lightpink}{rgb}{0.980, 0.941, 0.941}
\definecolor{lightyellow}{HTML}{FCFCE9}
\definecolor{lightblue}{HTML}{F2F7FE}

\usepackage[pagebackref,breaklinks,colorlinks,allcolors=cvprblue]{hyperref}

\title{GDPO-SR: Group Direct Preference Optimization for \\ One-Step Generative Image Super-Resolution}

\author{Qiaosi Yi$^{1,2}$,  Shuai Li$^{1}$,  Rongyuan Wu$^{1,2}$, Lingchen Sun$^{1,2}$, Zhengqiang Zhang$^{1,2}$, Lei Zhang$^{1,2}$\thanks{Corresponding author. This research is supported by the PolyU-OPPO Joint Innovative Research Center.} \\
{$^{1}$The Hong Kong Polytechnic University \qquad $^{2}$OPPO Research Institute} \\
{\tt\small  qiaosiyijoyies@gmail.com, cslzhang@comp.polyu.edu.hk}  \\ 
{\tt\small \{novak.li, rong-yuan.wu, ling-chen.sun, zhengqiang.zhang\}@connect.polyu.hk} 
}

\begin{document}
\maketitle
\begin{abstract}
Recently, reinforcement learning (RL) has been employed for improving generative image super-resolution (ISR) performance. However, the current efforts are focused on multi-step generative ISR, while one-step generative ISR remains underexplored due to its limited stochasticity. In addition, RL methods such as Direct Preference Optimization (DPO) require the generation of positive and negative sample pairs offline, leading to a limited number of samples, while Group Relative Policy Optimization (GRPO) only calculates the likelihood of the entire image, ignoring local details that are crucial for ISR. 
In this paper, we propose Group Direct Preference Optimization (GDPO), a novel approach to integrate RL into one-step generative ISR model training. First, we introduce a noise-aware one-step diffusion model that can generate diverse ISR outputs. To prevent performance degradation caused by noise injection, we introduce an unequal-timestep strategy to decouple the timestep of noise addition from that of diffusion. We then present the GDPO strategy, which integrates the principle of GRPO into DPO, to calculate the group-relative advantage of each online generated sample for model optimization. Meanwhile, an attribute-aware reward function is designed to dynamically evaluate the score of each sample based on its statistics of smooth and texture areas. Experiments demonstrate the effectiveness of GDPO in enhancing the performance of one-step generative ISR models. Code: https://github.com/Joyies/GDPO.
\end{abstract}    
\section{Introduction}
\label{sec:intro}
Different from classical image super-resolution (ISR) \cite{edsr,rcan,rdn,san,liang2021swinir,chen2025generalized}, which aims to reconstruct a high-resolution (HR) image from its low-resolution (LR) counterpart with known and relatively simple degradations (\eg, bicubic downsampling), real-world ISR (Real-ISR) \cite{srgan,wang2018esrgan,liang2022details,zhang2021designing,wang2021real,sun2024perception} aims to reconstruct HR images from LR inputs captured under real-world conditions, which are often corrupted with complex and unknown degradations. Real-ISR is more ill-posed than classical ISR due to the complex degradation, and the research focuses on how to synthesize realistic details without introducing many visual artifacts. While many GAN-based methods \cite{srgan,wang2018esrgan,zhang2021designing,wang2021real} have been proposed for Real-ISR, in recent years, diffusion models \cite{ho2020denoising, song2020score, dhariwal2021diffusion} have demonstrated significantly stronger capabilities in synthesizing more natural ISR images with richer details \cite{kawar2021snips, kawar2022denoising, wang2022zero, yue2023ResShift,wang2023exploiting,wu2023seesr,yang2023pixel,yi2025fine}.

\begin{figure*}
    \centering
    \includegraphics[width=0.8\linewidth]{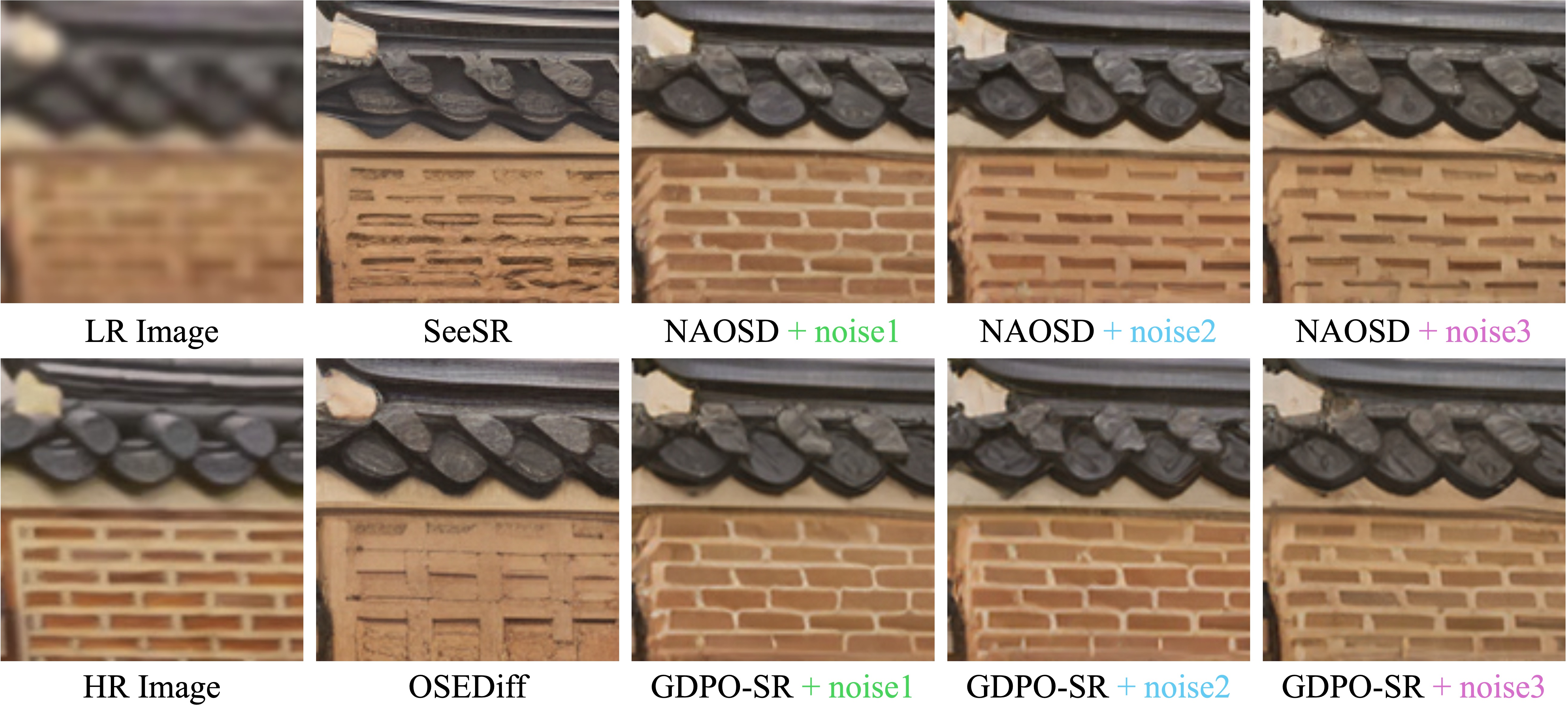}
    \vspace{-5mm}
    \caption{
    Noise regulates the diversity of generated samples. Different noise inputs yield both high-quality samples (\eg, noise1) and low-quality ones (\eg, noise2 and noise3). After preference learning, the model produces more visually pleasing results.
    }
    \label{fig:introfig1}
    \vspace{-5.5mm}
\end{figure*}

In particular, pre-trained large-scale text-to-image (T2I) diffusion models \cite{sd, flux2024} have been prevalently used as the backbones in Real-ISR tasks due to their powerful generative priors \cite{zhang2023adding,wang2023exploiting,wu2023seesr,wu2024one,zhang2024degradation,yue2025arbitrary,sun2024pixel,yi2025fine}. StableSR \cite{wang2023exploiting} firstly adapts the Stable Diffusion (SD) model \cite{sd} to Real-ISR, showcasing the potentials of pre-trained diffusion priors for enhancing the quality of LR images. PASD \cite{yang2023pixel} and SeeSR \cite{wu2023seesr} demonstrate that prompts can better activate the generative capacity of SD models, further improving the visual quality of Real-ISR output. These early SD-based methods employ the LR image as a control signal and start from noise with multi-step denoising to produce outputs, which incur significant computational cost and are prone to hallucination (see Fig.~\ref{fig:introfig1}). To accelerate inference speed, one-step diffusion-based Real-ISR methods \cite{wu2024one,zhang2024degradation,yue2025arbitrary,sun2024pixel,yi2025fine} have been proposed, which eliminate random noise initialization by directly taking the LR image as input. However, these methods suffer from limited generative capacity, sacrificing the details of Real-ISR results (see Fig.~\ref{fig:introfig1}). This motivates us to investigate whether we can find a new training paradigm to improve the generative capacity of one-step diffusion based Real-ISR models.

Recently, Reinforcement Learning (RL) techniques \cite{rafailov2023direct,wallace2024diffusion,shao2024deepseekmath,liu2025flow,wu2025visualquality}, especially those preference optimization algorithms such as Direct Preference Optimization (DPO) \cite{rafailov2023direct} and Group Relative Policy Optimization (GRPO) \cite{shao2024deepseekmath}, have demonstrated remarkable success in aligning model outputs with human preferences in various tasks, including large language models \cite{shao2024deepseekmath,yu2025dapo} and image generation \cite{wallace2024diffusion,liu2025flow,xue2025dancegrpo}, \etc. Inspired by this, the pioneering work DP$^2$OSR \cite{wu2025dp2osr} has successfully applied Diffusion-DPO \cite{wallace2024diffusion} to multi-step generative Real-ISR, reducing hallucinated details while improving visual quality. Therefore, one natural question arises: \textit{Can we leverage RL to enhance the performance of one-step generative Real-ISR models?}

Unfortunately, there are several challenges that hinder the application of DPO and GRPO to one-step Real-ISR models. Firstly, most of the existing one-step models \cite{wu2024one,zhang2024degradation,sun2024pixel,yi2025fine} directly map the LR image to its HR counterpart. Such a deterministic nature hinders the application of DPO and GRPO to Real-ISR, \ie, the policy model should be able to generate diverse outputs for the same input. Secondly, existing RL algorithms exhibit limitations when applied to Real-ISR tasks. DPO performs preference optimization using only a single pair of offline-generated positive and negative samples. This inevitably restricts data diversity and limits the model performance. GRPO alleviates this issue by generating multiple samples online and computing the group-relative advantage for each sample, which improves sample efficiency. However, GRPO only calculates the likelihood of the entire image, overlooking the local details of the image, which can degrade the visual quality of the reconstructed images.

To address these challenges, we propose Group Direct Preference Optimization (GDPO), which integrates the advantages of DPO and GRPO, for effective one-step generative super-resolution. First, we introduce a noise-aware one-step diffusion (NAOSD) as the base model to inject controllable noise into the latent features. By sampling different noises, the model generates outputs of varying quality from the same LR input. As shown in Fig.~\ref{fig:introfig1}, some noise inputs yield high-quality samples, while others produce suboptimal ones. This variability deterministic nature of existing one-step models and provides the diversity requirement for RL-based optimization. To avoid performance degradation caused by noise injection, we propose an unequal-timestep strategy to decouple the timestep of noise addition from that of diffusion denoising. Then, we present GDPO to combine the strengths of DPO and GRPO by training with online-generated groups of samples. 

GDPO consists of two core stages: advantage calculation and policy optimization. In the advantage calculation stage, to effectively distinguish the quality differences among samples within a group, we introduce an attribute-aware reward function (ARF) that adaptively balances fidelity-related and perception-related metrics. In the policy optimization stage, we reformulate the Diffusion-DPO loss \cite{wallace2024diffusion} to leverage group-relative advantages, prioritizing higher-reward samples while reducing the influence of less desirable ones, effectively combining the precision of DPO with the efficiency of GRPO.
As shown in Fig.~\ref{fig:introfig1}, our proposed GDPO-based Real-ISR model, GDPO-SR in short, reconstructs clear and regular brick textures. 

In summary, we introduce GDPO, a novel RL-based framework built upon a noise-aware one-step diffusion model, facilitating controllable output diversity and adaptive preference optimization for Real-ISR. Extensive experiments show that GDPO-SR achieves clearer and more detailed super-resolved images while reducing artifacts compared to existing methods.

\section{Related Work}
\label{sec:rw}

\textbf{Real-World Image Super-Resolution}.
Conventional ISR methods \cite{edsr,rcan,rdn,san,liang2021swinir,chen2025generalized, johnson2016perceptual,ssim,goodfellow2014generative,peng2025pixel} mainly focus on enhancing image fidelity by designing advanced network architectures and loss functions, whereas Real-ISR leverages generative priors to improve the perceptual quality of reconstructed images. Conventional Real-ISR approaches \cite{srgan,wang2018esrgan,zhang2021designing,wang2021real} mainly employ GAN \cite{goodfellow2014generative} to enhance perceptual realism, but suffer from training instability and artifacts \cite{liang2022details,sun2024perception}. Recently, diffusion models \cite{ho2020denoising, song2020score, dhariwal2021diffusion, zhang2023adding, peng2024towards} have inspired a surge of Real-ISR methods. Early attempts \cite{kawar2021snips, kawar2022denoising, wang2022zero, yue2023ResShift} train diffusion models from scratch; for example, ResShift \cite{yue2023ResShift} reformulates noise addition as residual shifting to fit Real-ISR task. However, these methods are limited in generative capacity. With the emergence of large-scale pre-trained T2I models such as SD \cite{sd} and FLUX \cite{flux2024}, researchers have begun to exploit their powerful generative priors for Real-ISR \cite{wang2023exploiting, wu2023seesr, yang2023pixel,duan2025dit4sr}. DiffBIR \cite{lin2023diffbir}, SeeSR \cite{wu2023seesr}, and PASD \cite{yang2023pixel} demonstrate that leveraging SD's priors through multi-step denoising can substantially improve the visual quality of reconstructed images. Nevertheless, these approaches suffer from high computational cost and hallucination of details. One-step diffusion-based methods \cite{wang2023sinsr,xie2024addsr,wu2024one,zhang2024degradation,yue2025arbitrary,sun2024pixel,yi2025fine,dong2025tsd,zhang2025time} adopt distillation losses \cite{wang2023sinsr,wang2023prolificdreamer,yu2023text,dong2025tsd} to transfer the generative capacity of multi-step models into a one-step framework, achieving visually pleasing results within a single step. However, the one-step constraint limits their generative capacity, motivating us to more effectively exploit the generative prior of diffusion models while maintaining efficiency.

\begin{figure*}
    \centering
    \includegraphics[width=0.9\linewidth]{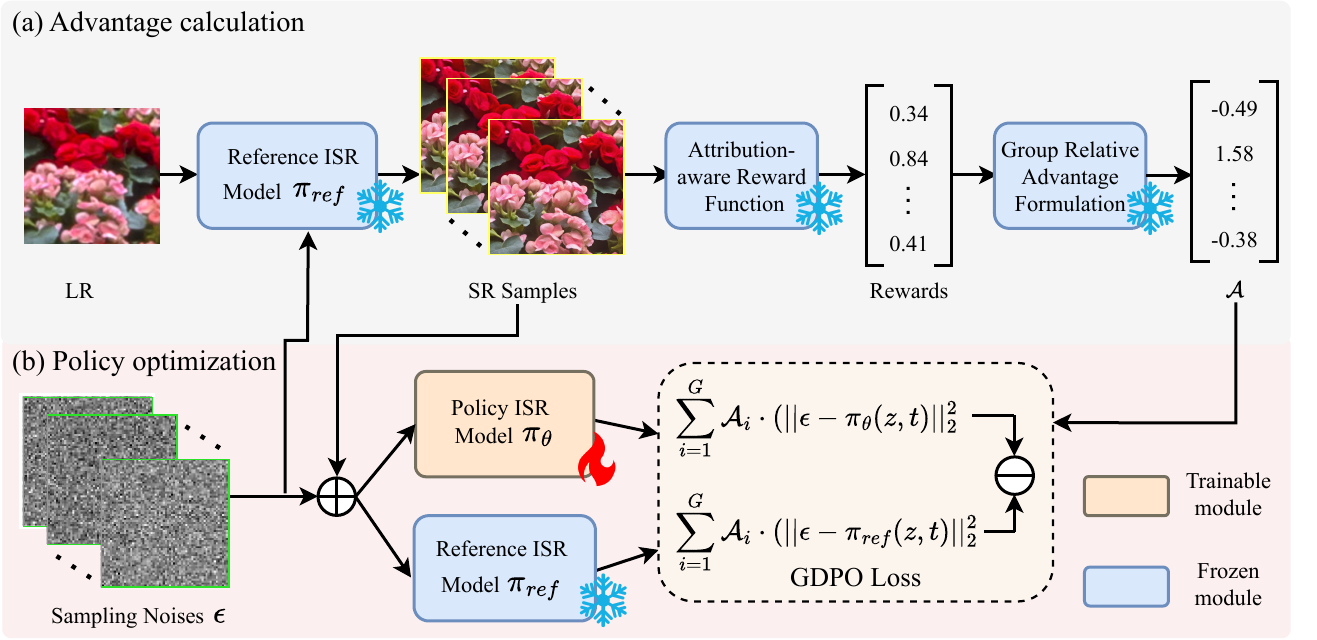}
    \vspace{-3mm}
    \caption{The framework of GDPO, which consists of two core stages: (a) advantage calculation and (b) policy optimization. Firstly, we employ a pre-trained one-step Real-ISR model as the reference model to generate a group of diverse outputs by injecting different random noises. Subsequently, we compute the advantage $\mathcal{A}$ for each sample by evaluating its reward with our designed attribute-aware reward functions and converting these rewards into group-relative advantages. In the policy optimization stage, we feed these samples along with noises into both the policy model and the reference ISR model, and update the parameters of the policy ISR model by minimizing the proposed GDPO loss, steering it to favor generating high-reward samples.
}
    \label{fig:model}
    \vspace{-3mm}
\end{figure*}

\noindent\textbf{Reinforcement Learning for Real-ISR}. 
Two representative reinforcement learning (RL) algorithms are DPO \cite{rafailov2023direct} and GRPO \cite{shao2024deepseekmath}. DPO employs offline-generated positive and negative pairs, encouraging the model to prefer the positive one. In contrast, GRPO performs online optimization by generating a group of samples and computing the relative advantage of each sample, guiding the model to favor those with higher advantages. Both DPO and GRPO have been widely adopted in image generative tasks \cite{wallace2024diffusion,liu2025flow}. Diffusion‑DPO \cite{wallace2024diffusion} extends DPO to diffusion models by reformulating the training objective with an evidence lower bound (ELBO), achieving stable preference alignment. Flow‑GRPO \cite{liu2025flow} and DanceGRPO \cite{xue2025dancegrpo} integrate GRPO into diffusion frameworks, updating models based on each sample’s relative advantage and likelihood. Inspired by these works, DP$^2$OSR \cite{wu2025dp2osr} applies Diffusion-DPO to multi‑step Real-ISR. However, Diffusion‑DPO relies solely on offline paired data, limiting sample diversity and hindering model performance. Meanwhile, Flow‑GRPO and DanceGRPO calculate the likelihood of the entire image, overlooking local details.

\section{Preliminaries}

\textbf{Direct Preference Optimization (DPO)}.
DPO \cite{rafailov2023direct} directly optimizes a generative policy by maximizing the likelihood of positive responses over negative ones. Diffusion-DPO \cite{wallace2024diffusion} extends the preference objective from intractable image-level likelihood to the more tractable reverse diffusion trajectory. It formulates a pixel-level constraint over the one-step denoising process, making preference optimization feasible and effective in the diffusion setting:
\begin{equation}
\begin{aligned}
L(\theta)
= - \mathrm{E}_{
(x^w_0, x^l_0) \sim \mathcal{D},
x^w_{t}\sim q(x^w_{t}|x^w_0),x^l_{t} \sim q(x^l_{t}|x^l_0)
} 
\log\sigma ( \\-\omega (
(\| \epsilon^w -\pi_\theta(x^w_{t},t)\|^2_2 - \|\epsilon^w - \pi_{ref}(x^w_{t},t)\|^2_2) \\
-( \| \epsilon^l -\pi_\theta(x^l_{t},t)\|^2_2 - \|\epsilon^l - \pi_{ref}(x^l_{t},t)\|^2_2))
),
\label{eq:loss-dpo-1}
\end{aligned}
\end{equation}
where $\pi_\theta$ and $\pi_{ref}$ are the policy and reference diffusion models, respectively, $\mathcal{D}$ is the dataset of preference pairs, ($x_0^{w}$, $x_0^{l}$) are the positive and negative images, $x_t=\sqrt{\alpha_t} x_0 + \sqrt{\beta_t} \epsilon$, $t$ is the timestep, $\epsilon$ is the random noise, $\alpha_t + \beta_t = 1$, $\omega$ is a hyperparameter, and $\sigma(\cdot)$ is the sigmoid function normalizing the preference score.

\noindent\textbf{Group Relative Policy Optimization (GRPO)}. GRPO \cite{shao2024deepseekmath} estimates the advantage of each sample by comparing the group-relative rewards of outputs generated by the policy model under the same input. 
Flow‑GRPO \cite{liu2025flow} and DanceGRPO \cite{xue2025dancegrpo} extend this mechanism to flow-matching models by converting the deterministic ODE sampling into an equivalent SDE form to introduce stochasticity, enabling us to compute the likelihood $p(x_t|t,c)$ of the entire image and measure the probability differences across policies:
\begin{equation}
\max_{p_{\theta}}
\mathbb{E}_{\{x^i_{0:T}\}_{i=1}^{G}
 \sim p_{\theta_{\text{old}}}(\cdot|c)}
\!\left[\sum_{i=1}^{G} \sum_{t=1}^{T}
\frac{
p_{\theta}(x^i_{t}|t, c)
}{
p_{\theta_{\text{old}}}(x^i_{t}|t, c)
} A_{i}
\right],
\label{eq:grpo}
\end{equation}
where $p_\theta$ and $p_{\theta_{old}}$ represent the likelihood of the policy model and old policy model, respectively. $G$ is the group size. $A_i$ is the group‑relative advantage for the $i$-th sample. $T$ is the number of denoising steps. $c$ is the text embedding. For simplicity, \cref{eq:grpo} omits the KL regularization term, the clip term, and the normalization factor $\frac{1}{GT}$, which are typically included in practice to stabilize training.

\section{Method}

We propose Group Direct Preference Optimization (GDPO), a novel RL-based training paradigm for one-step generative ISR. As illustrated in Fig. \ref{fig:model}, our framework consists of two key components: (1) a noise-aware one-step diffusion model as the policy model that generates diverse sample groups from the LR input, and (2) the preference optimization that updates the model based on relative advantages within sample groups. The objective of GDPO is to overcome the deterministic limitation of existing one-step Real-ISR methods, which inherently prevent the application of preference optimization algorithms, while simultaneously addressing the limitations of DPO and GRPO; that is, DPO relies on limited preference pairs and GRPO oversees local perceptual quality.

\begin{figure}
    \centering
    \includegraphics[width=1\linewidth]{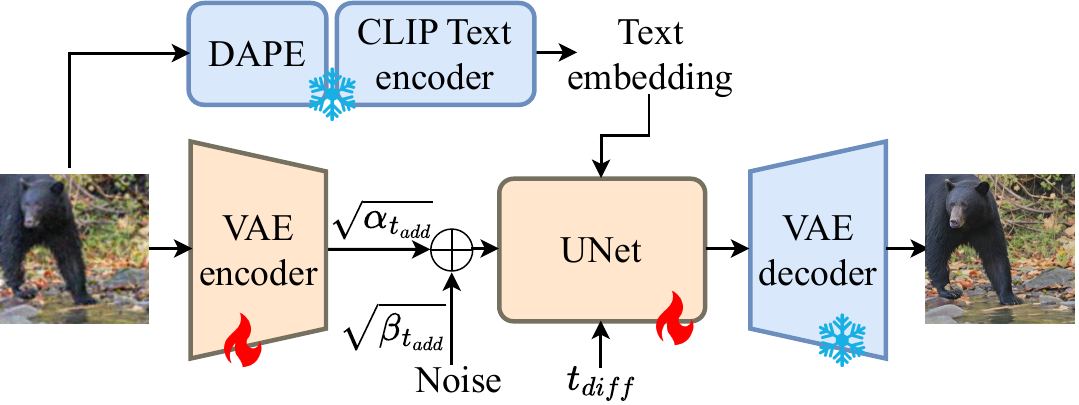}
    \vspace{-6mm}
    \caption{The structure of NAOSD, which uses the $t_{add}$ to control the intensity of injected noise.}
    \label{fig:noisemodel}
    \vspace{-3mm}
\end{figure}

\subsection{Noise-Aware One-Step Diffusion Model}

To enable diverse generation from deterministic one-step Real-ISR models, we design a noise-aware one-step diffusion (NAOSD) architecture that injects controllable noise into the latent space, as illustrated in Fig.~\ref{fig:noisemodel}. Given an LR input $I_{LR}$, we first obtain its latent feature $z_{LR} = E(I_{\mathrm{LR}})$ through a VAE encoder $E$. Meanwhile, we derive semantic guidance by extracting a text embedding $c_t$ using a composite prompt extraction module that integrates both DAPE~\cite{wu2023seesr} and the CLIP text encoder~\cite{rombach2022high}. To enable stochastic sampling, we inject controllable Gaussian noise $\epsilon$ into the latent $z_{LR}$:
\begin{equation}
\tilde{z} = \sqrt{\alpha_{t_{add}}}z_{LR} + \sqrt{\beta_{t_{add}}}\epsilon,\quad 
    \epsilon \sim \mathcal{N}(0,\mathbf{I}),
\label{eq:znoise}
\end{equation}
where $\alpha_{t_{add}}+\beta_{t_{add}}=1$. The perturbed latent $\tilde{z}$ is then denoised by the UNet at a diffusion timestep $t_{diff}$, producing the restored latent $z_{SR}$.
\begin{equation}
    z_{SR}=\frac{({\tilde{z}-\sqrt{\beta_{t_{diff}}}\,\,\text{UNet}(\tilde{z},c_t,t_{{diff}})})}{\sqrt{\alpha_{t_{diff}}}},
\label{eq:zsr}
\end{equation}
where $\alpha_{t_{diff}}+\beta_{t_{diff}}=1$. Finally, the VAE decoder $D$ maps the restored latent to the super-resolved image $I_{SR} = D\!\left(z_{{SR}}\right)$.
Following prior work \cite{wu2024one}, we employ Low-Rank Adaptation (LoRA) \cite{lora} to fine-tune both the VAE encoder and the UNet. The model is trained with a combination of $L_1$ loss, LPIPS loss and VSD loss:
\begin{equation}
\begin{aligned}
    \mathcal{L}_{onestep} & = L_1(I_{SR}, I_{HR}) + \lambda_{1}L_{LPIPS}(I_{SR}, I_{HR}) \\ 
    & + \lambda_{2}L_{VSD}(I_{SR}, I_{HR}),
\end{aligned}
\end{equation}
where $\lambda_{1}=2$, $\lambda_{2}=1$ are weighting hyper-parameters.

In the above noise-injection and one-step denoising framework, $t_{add}$ determines the injected noise strength and the upper bound of diversity, while $t_{diff}$ governs the denoising strength and reconstruction fidelity. Empirically, if we set $t_{add}$ = $t_{diff}$ and increase the timestep of them, the generative capability of the model improves but the fidelity degrades notably. To balance this trade-off, we propose an unequal-timestep strategy, which set a larger $t_{add}$ to expand the sampling space while adopting a more conservative $t_{diff}$ to stabilize fidelity.

\noindent\textbf{Remark} (Diversity induced by the noise).
We provide an approximate analysis to show that noise injection preserves a residual stochastic term, enabling diverse outputs. Assuming perfect noise prediction ($\text{UNet}(\tilde{z},c_t,t_{{diff}})\!\approx\!\epsilon$), substituting \cref{eq:znoise} into \cref{eq:zsr} gives the approximate solution: 
\begin{equation}
z_{SR}\!\approx\!\frac{\sqrt{\alpha_{t_{add}}}}{\sqrt{\alpha_{t_{diff}}}}z_{LR}\!+\!\frac{\sqrt{\beta_{t_{add}}}-\sqrt{\beta_{t_{diff}}}}{\sqrt{\beta_{t_{diff}}}}\epsilon. 
\end{equation}
This approximation indicates that when $t_{add} \neq t_{diff}$, an additional noise term $\epsilon$ is introduced, increasing diversity. 

\subsection{Group Direct Preference Optimization}

GDPO utilizes a group of online generated samples to optimize the policy model. As illustrated in Fig. \ref{fig:model}, GDPO operates in two sequential stages: \textit{Advantage Calculation}, which computes the relative advantages $\mathcal{A}$ of a group of online generated ISR samples, and \textit{Policy Optimization}, which optimizes the policy model to favor higher-advantage samples. Note that in GDPO, both the reference ISR model $\pi_{ref}$ and the policy ISR model $\pi_{\theta}$ are initialized with the pretrained NAOSD model.

\noindent \textbf{Advantage Calculation.} For an LR image $I_{LR}$, we generate a group of $G$ ISR candidates $\mathcal{S}=\{I^{i}_{SR}\}^G_{i=1}$ by injecting different random noise $\{\epsilon_i\}^G_{i=1}$ into the model. Subsequently, we design an attribute-aware reward function (ARF) to assess each generated sample. Similar to the reward function in DP$^2$O-SR \cite{wu2025dp2osr}, our ARF combines full-reference (FR) metrics ($\mathcal{G}_{FR}$) and no-reference (NR) metrics ($\mathcal{G}_{NR}$): the former measure the fidelity to the HR image, while the latter capture perceptual quality. Specifically, in our ARF, $\mathcal{G}_{FR}$ only includes PSNR due to its strong capability to measure image fidelity, while $\mathcal{G}_{NR}$ includes MANIQA~\cite{maniqa} and MUSIQ~\cite{musiq} as they are widely used for perceptual quality evaluation.

\begin{figure}
    \centering
    \includegraphics[width=0.9\linewidth]{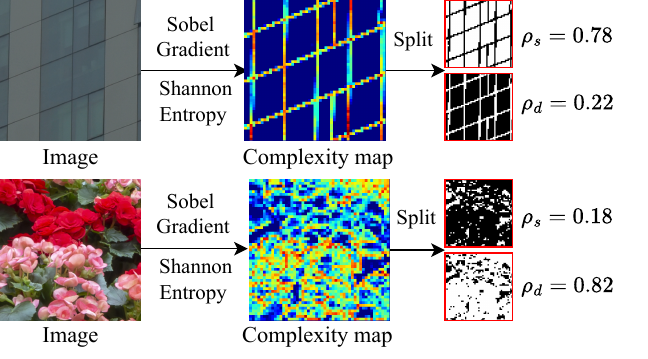}
    \vspace{-4mm}
    \caption{The pipeline of calculating smooth and detailed regions.}
    \label{fig:smoothdetails}
    \vspace{-3.5mm}
\end{figure}

In addition, different images demand different balances between fidelity and perception—for example, building scenes often favor fidelity, whereas foliage or flower scenes may prioritize perceptual attractiveness. Therefore, we dynamically adjust their weights according to the proportion of smooth and detailed regions in the image. Specifically, we first convert the image to grayscale and partition it into $10\times10$ patches. For each patch, we compute the Sobel gradient magnitude, build its histogram, and compute the Shannon entropy to measure the complexity of patch details. According to the complexity map, we split the image into smooth regions (low complexity), denoted by $\Omega_s$, and detailed regions (high complexity), denoted by $\Omega_d$, and use their proportions to adaptively weight the reference-based and no-reference rewards:
\begin{equation}
    R_{i}=\rho_s \sum_{f \in \mathcal{G}_{FR}} \frac{s^f_i}{|\mathcal{G}_{FR}|} +\rho_d \sum_{f \in \mathcal{G}_{NR}} \frac{s^f_i}{|\mathcal{G}_{NR}|}, i \in [1:G]
    \label{eq:reward}
\end{equation}
where $s^f_i$ is the min–max normalized score of a metric $f$ (the higher the better). $|\mathcal{G}_{FR}|$ and $|\mathcal{G}_{NR}|$ are the numbers of metrics in $\mathcal{G}_{FR}$ and $\mathcal{G}_{NR}$, respectively. $\rho_{s} = \frac{|\Omega_{s}|}{(|\Omega_s|+|\Omega_d|)}$ and $\rho_{d} = \frac{|\Omega_{d}|}{(|\Omega_s|+|\Omega_d|)}$ denote the proportions of smooth and detailed regions, respectively, where $|\Omega_s|(|\Omega_d|)$ is the number of pixels in the smooth (detailed) region. Fig. \ref{fig:smoothdetails} shows the calculation process of smooth and detailed regions.

Reward scores from \cref{eq:reward} reflect each sample’s absolute reward but not their relative reward. We further employ the group relative advantage formulation as in GRPO \cite{shao2024deepseekmath} to compute each sample's group-relative advantage $\mathcal{A}_i$, which can inform the model that which candidates are better or worse within the group:
\begin{equation}
    \mathcal{A}_i=\frac{R_i-\mathrm{mean}(\{R_j\}_{j=1}^G)}{\mathrm{std}(\{R_j\}_{j=1}^G)}.
\end{equation}

\noindent \textbf{Policy Optimization.} 
Unlike DPO, which only uses one paired sample, GDPO uses a set of generated samples and the calculated relative advantage $\mathcal{A}$ to update the policy model. The GDPO loss is computed as follows:
\begin{equation}
\begin{aligned}
L_{GDPO}
= - \mathrm{E}_{
x_0 \sim \mathcal{D},
x_{t}\sim q(x_{t}|x_0)
} 
\log\sigma (-\omega ( \,\,\,\,\,\,\,\,\,\,\,\,\,\,\,\,\,\,\,\,\,\,\,\\
\sum^G_{i=1} \mathcal{A}_i(\| \epsilon -\pi_\theta(x_{t},t)\|^2_2 - \|\epsilon - \pi_{ref}(x_{t},t)\|^2_2))
),\label{eq:gdpo}
\end{aligned}
\end{equation}
where $x_t$ is the noisy latent of ISR candidates. We can see that DPO is a special case of our GDPO with $G=2$. GDPO favors candidates with larger advantages: when sample $i$ has a higher reward, its $\mathcal{A}_i$ increases and carries more weight in $\sum^G_{i=1} \mathcal{A}_i(\cdot)$, making $-\omega\sum^G_{i=1} \mathcal{A}_i(\cdot)$ more likely to be negative, which raises $\log\sigma(\cdot)$ and drives the gradient to prioritize these high-reward samples—achieving alignment biased toward the higher-reward side. In contrast to GRPO, which requires computing the full-image likelihood, GDPO inherits Diffusion-DPO’s advantage of implicit likelihood computation while imposing pixel-level constraints, thereby learning local details more effectively.
\section{Experiment}
\begin{table*}[!tb]
\caption{Performance comparison with the base model NAOSD on real-world and synthetic datasets. Metrics with a {\sethlcolor{lightblue}\hl{blue background}} denote those utilized in the reward function, while {\sethlcolor{lightyellow}\hl{yellow-shaded}} ones correspond to metrics that are excluded from it. Arrows denote if higher (↑) or lower (↓) values represent better performance. The best results are highlighted in {\color[HTML]{FF0000}\textbf{red}}.}
\vspace{-3mm}
\centering
	\resizebox{0.9\linewidth}{!}{
\begin{tabular}{c|c|
>{\columncolor[HTML]{F2F7FE}}c 
>{\columncolor[HTML]{FCFCE9}}c 
>{\columncolor[HTML]{FCFCE9}}c 
>{\columncolor[HTML]{FCFCE9}}c 
>{\columncolor[HTML]{FCFCE9}}c 
>{\columncolor[HTML]{F2F7FE}}c 
>{\columncolor[HTML]{F2F7FE}}c 
>{\columncolor[HTML]{FCFCE9}}c 
>{\columncolor[HTML]{FCFCE9}}c }
\toprule
Dataset                   & Method    & PSNR↑                                 & SSIM↑                                   & LPIPS↓                                  & FID↓                           & DISTS↓                         & MANIQA↑                        & MUSIQ↑                        & CLIPIQA↑                       & AFINE↓                        \\ \hline
\hhline{-----------} 
                          & NAOSD & {\color[HTML]{333333} 27.97}          & {\color[HTML]{333333} 0.7831}          & {\color[HTML]{333333} 0.2859}          & 140.37 & {\color[HTML]{333333} 0.2113} & {\color[HTML]{333333} 0.6041} & {\color[HTML]{333333} 65.17} & {\color[HTML]{333333} 0.6886} & {\color[HTML]{333333} 20.52} \\
\multirow{-2}{*}{DrealSR} & GDPO-SR & {\color[HTML]{FF0000}\textbf{ 28.18}} & {\color[HTML]{FF0000}\textbf{ 0.7839}} & {\color[HTML]{FF0000}\textbf{ 0.2851}} & {\color[HTML]{FF0000}\textbf{ 138.87}} & {\color[HTML]{FF0000}\textbf{ 0.2112}} & {\color[HTML]{FF0000}\textbf{ 0.6180}} & {\color[HTML]{FF0000}\textbf{ 65.63}} & {\color[HTML]{FF0000}\textbf{ 0.7020}} & {\color[HTML]{FF0000}\textbf{ 18.72}} \\ \hline
\hhline{-----------} 
                          & NAOSD & {\color[HTML]{333333} 25.25}          & {\color[HTML]{FF0000}\textbf{ 0.7346}}          & {\color[HTML]{333333} 0.2689}          & {\color[HTML]{333333} 114.91} & {\color[HTML]{333333} 0.2001} & {\color[HTML]{333333} 0.6459} & {\color[HTML]{333333} 69.06} & {\color[HTML]{333333} 0.6617} & {\color[HTML]{333333} 18.90} \\
\multirow{-2}{*}{RealSR}  & GDPO-SR & {\color[HTML]{FF0000}\textbf{ 25.48}}          & {\color[HTML]{333333} 0.7328}          & {\color[HTML]{FF0000}\textbf{ 0.2675} }         & {\color[HTML]{FF0000}\textbf{ 112.13}} & {\color[HTML]{FF0000}\textbf{ 0.1980}} & {\color[HTML]{FF0000}\textbf{ 0.6615}} & {\color[HTML]{FF0000}\textbf{ 69.42}} & {\color[HTML]{FF0000}\textbf{ 0.6760}} & {\color[HTML]{FF0000}\textbf{ 17.73}} \\
\hline
\hhline{-----------}  
                          & NAOSD & {\color[HTML]{333333} 23.83}          & {\color[HTML]{FF0000}\textbf{ 0.6131}}          & {\color[HTML]{FF0000}\textbf{ 0.2848}}          & {\color[HTML]{FF0000}\textbf{ 25.16}}  & {\color[HTML]{FF0000}\textbf{ 0.1922}} & {\color[HTML]{333333} 0.6297} & {\color[HTML]{333333} 68.20} & {\color[HTML]{333333} 0.6758} & {\color[HTML]{333333} 41.76} \\
\multirow{-2}{*}{DIV2K-val}   & GDPO-SR      & {\color[HTML]{FF0000}\textbf{ 23.92} }                               & {\color[HTML]{333333} 0.6117}          & {\color[HTML]{333333} 0.2897}          & {\color[HTML]{333333} 26.44}  & {\color[HTML]{333333} 0.1965} & {\color[HTML]{FF0000}\textbf{ 0.6423}} & {\color[HTML]{FF0000}\textbf{ 68.80}} & {\color[HTML]{FF0000}\textbf{ 0.6929}} & {\color[HTML]{FF0000}\textbf{ 41.56}} \\ \bottomrule
\end{tabular}}
\label{tab:gdporesult}
\vspace{-2mm}
\end{table*}

\begin{table*}[!tb]
\caption{Quantitative comparison with different methods on real-world and synthetic datasets. The best and second best results are highlighted in {\color[HTML]{FF0000} \textbf{red}} and {\color[HTML]{6434FC} \textbf{blue}}, respectively. Arrows denote if higher (↑) or lower (↓) values represent better performance.}
\vspace{-3mm}
\centering
	\resizebox{0.9\linewidth}{!}{
\begin{tabular}{c|c|ccccccccc }
\toprule
Dataset                   & Method    & PSNR↑                                 & SSIM↑                                   & LPIPS↓                                  & FID↓                           & DISTS↓                         & MANIQA↑                        & MUSIQ↑                        & CLIPIQA↑                       & AFINE↓                        \\ \hline
% \hline
% \hline
                          & StableSR  & 28.03                                 & 0.7536                                 & 0.3284                                 & 148.98                        & 0.2269                        & 0.5592                        & 58.51                        & 0.6356                        & 35.77                        \\
                          & DiffBIR   & 26.71                                 & 0.6571                                 & 0.4557                                 & 166.79                        & 0.2748                        & 0.5927                        & 61.07                        & 0.6395                        & 39.42                        \\
                          & SeeSR     & {\color[HTML]{6434FC} \textbf{28.07}}                                 & 0.7684                                 & 0.3174                                 & 147.39                        & 0.2315                        & 0.6054                        & {\color[HTML]{6434FC} \textbf{65.08}}                        & 0.6905                        & 22.79                        \\
                          & PASD      & 27.36                                 & 0.7073                                 & 0.3760                                 & 156.13                        & 0.2531                        & 0.6160                        & 64.87                        & 0.6808                        & 33.89                        \\
                          & OSEDiff   & 27.92                                 & {\color[HTML]{6434FC} \textbf{0.7835}}                                 & {\color[HTML]{6434FC} \textbf{0.2968}}                                 & {\color[HTML]{FF0000} \textbf{135.30}}                        & {\color[HTML]{6434FC} \textbf{0.2165}}                        & 0.5899                        & 64.65                        & {\color[HTML]{6434FC} \textbf{0.6963}}                        & {\color[HTML]{6434FC} \textbf{21.39}}                        \\
                          & InvSR     & 25.79                                 & 0.7176                                 & 0.3471                                 & 166.42                        & 0.2381                        & {\color[HTML]{FF0000} \textbf{0.6212}}                        & 64.92                        & 0.6918                        & 21.69                        \\ 
\rowcolor{lightpink} \multirow{-7}{*}{DrealSR} & GDPO-SR      & {\color[HTML]{FF0000} \textbf{28.18}} & {\color[HTML]{FF0000} \textbf{0.7839}} & {\color[HTML]{FF0000} \textbf{0.2851}} & {\color[HTML]{6434FC} \textbf{138.87}} & {\color[HTML]{FF0000} \textbf{0.2112}} & {\color[HTML]{6434FC} \textbf{0.6180}} & {\color[HTML]{FF0000} \textbf{65.63}} & {\color[HTML]{FF0000} \textbf{0.7020}} & {\color[HTML]{FF0000} \textbf{18.72}} \\ \hline
% \hline
                          & StableSR  & 24.64                                 & 0.7080                                 & 0.3002                                 & 128.51                        & 0.2140                        & 0.6215                        & 65.88                        & 0.6234                        & 27.62                        \\
                          & DiffBIR   & 24.75                                 & 0.6567                                 & 0.3636                                 & 128.99                        & 0.2312                        & 0.6252                        & 64.98                        & 0.6463                        & 32.64                        \\
                          & SeeSR     & 25.15                                 & 0.7210                                 & 0.3007                                 & 125.45                        & 0.2224                        & 0.6441                        & {\color[HTML]{FF0000} \textbf{69.82}}                        & 0.6707                        & 24.07                        \\
                          & PASD      & {\color[HTML]{6434FC} \textbf{25.21}}                                 & 0.6798                                 & 0.3380                                 & {\color[HTML]{6434FC} \textbf{124.28}}                        & 0.2260                        & 0.6493                        & 68.75                        & 0.6620                        & 39.00                        \\
                          & OSEDiff   & 25.15                                 & {\color[HTML]{FF0000} \textbf{0.7341}}                                 & 0.2921                                 & 123.49                        & 0.2128                        & 0.6326                        & 69.09                        & 0.6693                        & 20.92                        \\
                          & InvSR     & 24.30                                 & 0.7145                                 & {\color[HTML]{6434FC} \textbf{0.2775}}                                 & 129.52                        & {\color[HTML]{6434FC} \textbf{0.2060}}                        & {\color[HTML]{6434FC} \textbf{0.6561}}                       & 67.31                        & {\color[HTML]{6434FC} \textbf{0.6739}}                        & {\color[HTML]{FF0000} \textbf{16.58}}                        \\ 
\rowcolor{lightpink} \multirow{-7}{*}{RealSR}  & GDPO-SR      & {\color[HTML]{FF0000} \textbf{25.48}}          & {\color[HTML]{6434FC} \textbf{0.7328}}          & {\color[HTML]{FF0000} \textbf{0.2675}}          & {\color[HTML]{FF0000} \textbf{112.13}} & {\color[HTML]{FF0000} \textbf{0.1980}} & {\color[HTML]{FF0000} \textbf{0.6615}} & {\color[HTML]{6434FC} \textbf{69.42}} & {\color[HTML]{FF0000} \textbf{0.6760}} & {\color[HTML]{6434FC} \textbf{17.73}} \\ \hline
% \hline
                          & StableSR  & 23.26                                 & 0.5726                                 & 0.3113                                 & {\color[HTML]{FF0000} \textbf{24.44}}                         & 0.2048                        & 0.6190                        & 65.92                        & 0.6771                        & 49.31                        \\
                          & DiffBIR   & 23.64                                 & 0.5647                                 & 0.3524                                 & 30.73                         & 0.2128                        & 0.6209                        & 65.81                        & 0.6704                        & 49.58                        \\
                          & SeeSR     & {\color[HTML]{6434FC} \textbf{23.82}}                                 & {\color[HTML]{6434FC} \textbf{0.6096}}                                 & 0.3160                                 & {\color[HTML]{6434FC} \textbf{25.42}}                         & 0.1971                        & 0.6187                        & {\color[HTML]{6434FC} \textbf{68.50}}                        & 0.6893                        & 42.18                        \\
                          & PASD      & 23.23                                 & 0.5877                                 & 0.3634                                 & 30.12                         & 0.2200                        & {\color[HTML]{6434FC} \textbf{0.6420}}                        & 68.41                        & 0.6802                        & 45.19                        \\
                          & OSEDiff   & 23.72                                 & 0.6108                                 & {\color[HTML]{6434FC} \textbf{0.2941}}                                 & 26.32                         & {\color[HTML]{6434FC} \textbf{0.1976}}                        & 0.6148                        & 67.97                        & 0.6683                        & 44.69                        \\
                          & InvSR     & 23.10                                 & 0.5985                                 & 0.3045                                 & 28.45                         & 0.1985                        & 0.6406                        & 68.43                        & {\color[HTML]{FF0000} \textbf{0.7118}}                        & {\color[HTML]{FF0000} \textbf{37.51}}                        \\ 
\rowcolor{lightpink} \multirow{-7}{*}{DIV2K-val}     & GDPO-SR      & {\color[HTML]{FF0000} \textbf{23.92}}                                & {\color[HTML]{FF0000} \textbf{0.6117}}          & {\color[HTML]{FF0000} \textbf{0.2897}}          & 26.44  & {\color[HTML]{FF0000} \textbf{0.1965}} & {\color[HTML]{FF0000} \textbf{0.6423}} & {\color[HTML]{FF0000} \textbf{68.80}} & {\color[HTML]{6434FC} \textbf{0.6929}} & {\color[HTML]{6434FC} \textbf{41.56}} \\ \bottomrule
\end{tabular}}
\label{tab:sotaresult}
\vspace{-4mm}
\end{table*}

\subsection{Experimental Settings} 

\textbf{Training Datasets.} Similar to previous works \cite{wang2023exploiting,wu2024one}, our base model NAOSD is pre-trained on 1.5 million $512\times512$ image patches cropped from the LSDIR dataset \cite{li2023lsdir} and the first 10K face images of the FFHQ dataset \cite{ffhq}. For the GDPO stage, we further select 120,000 high-quality $512\times512$ image patches from LSDIR for fine-tuning. The Real-ESRGAN degradation pipeline \cite{wang2021real} is utilized to construct the low- and high-resolution training pairs.

\noindent\textbf{Testing Datasets.} To assess the effectiveness of our proposed GDPO, we adopt the benchmark datasets used in previous works \cite{wang2023exploiting,wu2024one,yi2025fine}, including DRealSR~\cite{drealsr}, RealSR~\cite{realsr} and DIV2K-val~\cite{div2k}. Specifically, RealSR and DRealSR contain 100 and 93 pairs of real-world low- and high-resolution images, respectively, where the LR and HR images are center-cropped to $128\times128$ and $512\times512$. In contrast, DIV2K-val consists of 3,000 synthetic LR-HR pairs of $512\times512$ resolution, generated using the Real-ESRGAN degradation pipeline. 

\noindent\textbf{Implementation Details.} The SD2.1-base \cite{rombach2022high} is used as the pre-trained diffusion model. In the pre-training of our base model NAOSD, we set batch size as 16 and run for 35,000 iterations on 4 NVIDIA A100 GPUs. The AdamW optimizer \cite{adamw} is used with the the rank of LoRA set to 4 and the learning rate set to $5\times10^{-5}$. In the GDPO fine-tuning stage, the LoRA rank is also set to 4. The group size of GDPO is set to 6, while the learning rate and batch size are configured to $5\times10^{-5}$ and 8, respectively. The preference weighting hyperparameter $\omega$ in \cref{eq:gdpo} is set to 5,000 and the GDPO-SR model is trained for 1,500 iterations on 8 NVIDIA A100 GPUs. 

\noindent\textbf{Evaluation Metrics.} The performance of the proposed model is evaluated across a set of widely used image quality assessment metrics, which are divided into two categories. (1) Metrics included in the ARF, comprising both FR metric PSNR and NR metrics MUSIQ~\cite{musiq} and MANIQA~\cite{maniqa}. (2) Metrics not included in the ARF, including FR metrics SSIM~\cite{ssim}, LPIPS~\cite{lpips}, FID~\cite{fid} and DISTS~\cite{dists}, and NR metrics CLIPIQA~\cite{clipiqa} and AFINE~\cite{wu2023q}.

\noindent\textbf{Compared Methods.} To demonstrate the effectiveness of GDPO-SR, we conduct comparisons against: (1) recent SD-based multi-step Real-ISR methods, including StableSR~\cite{wang2023exploiting}, PASD~\cite{yang2023pixel}, DiffBIR~\cite{lin2023diffbir}, and SeeSR~\cite{wu2023seesr}; (2) one-step diffusion approaches such as OSEDiff~\cite{wu2024one} and InvSR~\cite{yue2025arbitrary}; and (3) the recently developed RL-based Real-ISR method DP$^2$OSR \cite{wu2025dp2osr}.

\subsection{Experimental Results}

\begin{table*}[!tb]
\caption{Quantitative comparison with DP$^2$O-SR on real-world datasets. The best results are highlighted in {\color[HTML]{FF0000} red}.}
\vspace{-3mm}
\centering
	\resizebox{0.9\linewidth}{!}{
\begin{tabular}{c|c|ccccccccc }
\toprule
Dataset                   & Method    & PSNR↑                                 & SSIM↑                                   & LPIPS↓                                  & FID↓                           & DISTS↓                         & MANIQA↑                        & MUSIQ↑                        & CLIPIQA↑                       & AFINE↓                        \\ \hline \hhline{-----------}
& DP$^2$O-SR     & 23.95                                 & 0.6294                                 & 0.4412                                 & 157.76                        & 0.2763                       & {\color[HTML]{FF0000}0.6674}                       & {\color[HTML]{FF0000}71.09}                        & {\color[HTML]{FF0000}0.7842}                        & {\color[HTML]{FF0000}11.39}             \\ 
 \multirow{-2}{*}{DrealSR} & GDPO-SR      & {\color[HTML]{FF0000} 28.18} & {\color[HTML]{FF0000} 0.7839} & {\color[HTML]{FF0000} 0.2851} & {\color[HTML]{FF0000} 138.87} & {\color[HTML]{FF0000} 0.2112} & 0.6180 & 65.63 & 0.7020 & 18.72 \\ \hline \hhline{-----------}
                          & DP$^2$O-SR     & 22.39                                 & 0.6467                                 & 0.3513                                 & 131.79                        & 0.2442                       & {\color[HTML]{FF0000} 0.7036}                     & {\color[HTML]{FF0000} 73.24}                        & {\color[HTML]{FF0000} 0.7484}                        & {\color[HTML]{FF0000} 8.654}                       \\ 
 \multirow{-2}{*}{RealSR}  & GDPO-SR      & {\color[HTML]{FF0000} 25.48}          & {\color[HTML]{FF0000} 0.7328}          & {\color[HTML]{FF0000} 0.2675}          & {\color[HTML]{FF0000} 112.13} & {\color[HTML]{FF0000} 0.1980} & 0.6615 & 69.42 & 0.6760 & 17.73 \\ 
 \bottomrule
\end{tabular}}
\label{tab:rlresults}
\vspace{-3mm}
\end{table*}

\begin{figure*}[t]
    \centering
    \includegraphics[width=0.9\linewidth]{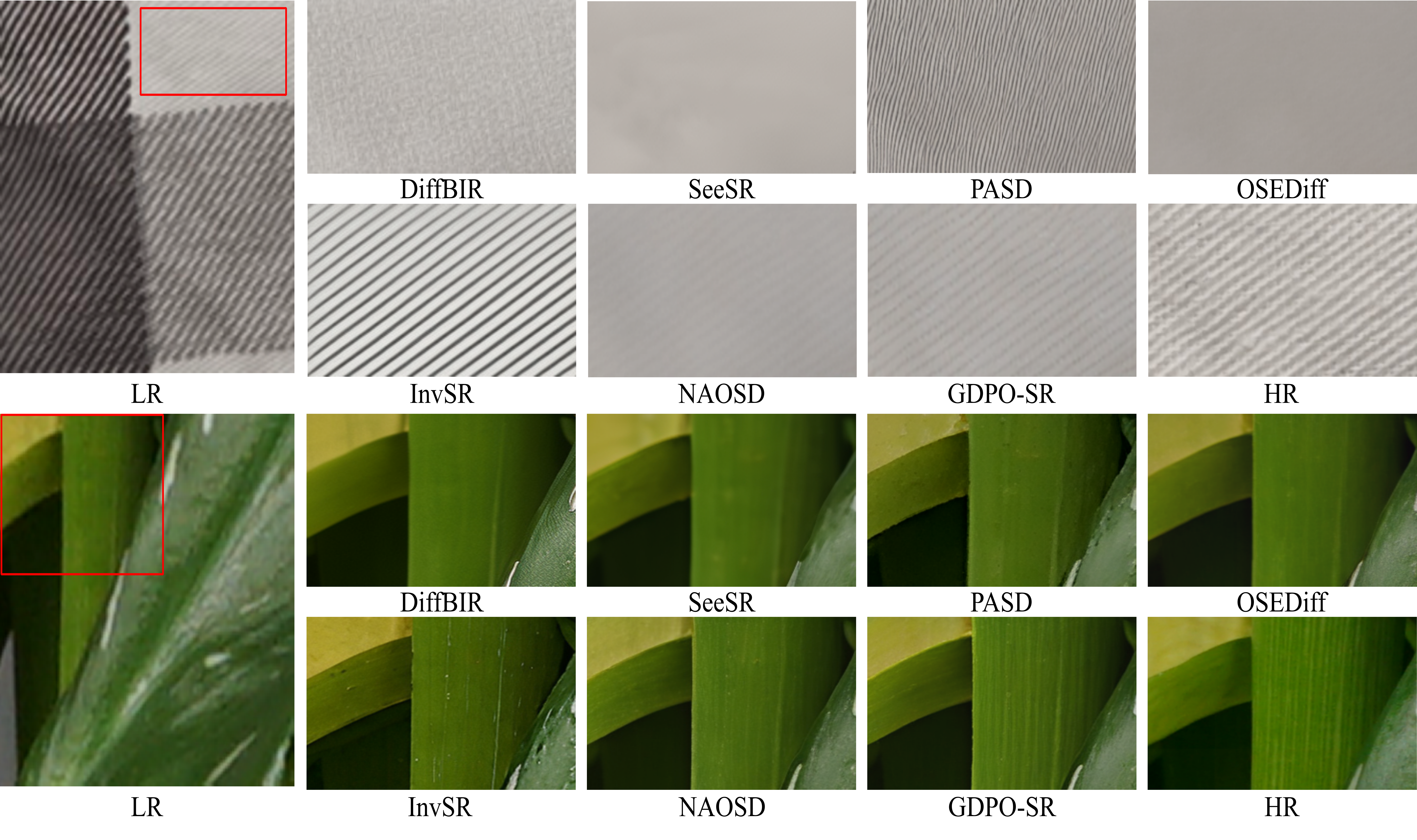}
    \vspace{-4mm}
    \caption{Visual comparison with SD-based Real-ISR methods. Please zoom in for a better view.}
    \label{fig:visualresults}
    \vspace{-4mm}
\end{figure*}

\noindent\textbf{Comparison with Base Model.} Table \ref{tab:gdporesult} presents a performance comparison between the proposed GDPO‑SR and its base model NAOSD. For fairness, NAOSD and GDPO‑SR are fed with identical noise for the same input. First, we can see that GDPO‑SR consistently achieves better results on all metrics involved in ARF. For instance, on the RealSR dataset, the PSNR increases from 25.25dB to 25.48dB, while MANIQA improves from 0.6459 to 0.6615, and MUSIQ rises from 69.06 to 69.42. 
Second, for metrics not included in the ARF, GDPO‑SR still outperforms NAOSD, particularly on NR metrics (CLIPIQA and AFINE). On the synthetic DIV2K‑val dataset, although GDPO‑SR does not surpass NAOSD in SSIM, LPIPS, FID and DISTS, it achieves improvements in PSNR and all NR metrics. In particular, GDPO-SR's advantages on both the real‑world datasets (DRealSR and RealSR) highlight its generalization capability under realistic degradations. Overall, GDPO‑SR effectively improves the performance of NAOSD, validating the effectiveness of the proposed GDPO strategy.

\noindent\textbf{Comparison with State-of-the-Arts.} Table \ref{tab:sotaresult} presents a quantitative comparison between GDPO‑SR and state‑of‑the‑art SD‑based Real‑ISR methods on real-world and synthetic datasets. For FR metrics, GDPO‑SR achieves the best performance among all methods on PSNR, LPIPS and DISTS across all the three datasets. GDPO‑SR also shows leading performance in other FR metrics, such as SSIM and FID, further demonstrating its strong ability to reproduce the fidelity and structure of the images.  
For NR metrics, GDPO‑SR still maintains overall leading performance, confirming that the model not only restores high‑fidelity details but also produces visually more natural and pleasing results. Specifically, on the DRealSR dataset, GDPO‑SR achieves the highest PSNR (28.18dB), SSIM (0.7839), MUSIQ (65.63), and CLIPIQA (0.7020), while showing the lowest LPIPS (0.2851) and DISTS (0.2112), significantly surpassing both multi‑step approaches (\eg, PASD, SeeSR) and one‑step methods (\eg, OSEDiff, InvSR). These results clearly validate GDPO‑SR's superior reconstruction quality and perceptual realism. 

\begin{table}
\caption{Comparison of mode size, running-time and FLOPs.}
 \vspace{-3mm}
	\centering
	\resizebox{\linewidth}{!}{
		\begin{tabular}{c|cccccc>{\columncolor{lightpink}}c}
			\toprule
			Methods & StableSR & DiffBIR & SeeSR & PASD &  OSEDiff & InvSR & GDPO-SR \\ \midrule
        Para.(B)   & 1.56     & 1.68    & 2.51  & 2.31  & 1.77   & 1.33 & 1.77 \\ 
Time(s)    & 10.03    & 2.72    & 4.30  & 2.80  & 0.11   & 0.12 & 0.11 \\ 
FLOPs(T)   & 79.94    & 24.31   & 65.86 & 29.13 & 2.27   & 2.40 & 2.27 \\
	\bottomrule
	\end{tabular}}
 \label{tab:flops}
 \vspace{-5mm}
\end{table}

\noindent\textbf{Qualitative Comparisons.} The visual comparisons are shown in Fig. \ref{fig:visualresults}. One can see that, compared with the base model NAOSD, GDPO-SR produces clearer structures and richer fine details. Compared with other SD-based methods, our GDPO-SR shows clear superiority. In the first example of Fig. \ref{fig:visualresults}, GDPO-SR reconstructs fine texture patterns that are more consistent with the HR image, while InvSR produces excessively saturated textures that deviate from the HR image. Similarly, in the second example, GDPO-SR can reconstruct clearer and more realistic plant textures and vein details, while competing methods produce over-smoothed results that lack realistic textures. These qualitative results highlight that GDPO-SR not only enhances image clarity but also preserves realistic and detailed structures, outperforming both the base model and other comparison methods. Due to page limit, more visual comparisons can be found in the \textbf{Supplementary Materials}.

\begin{figure}
    \centering
    \includegraphics[width=0.87\linewidth]{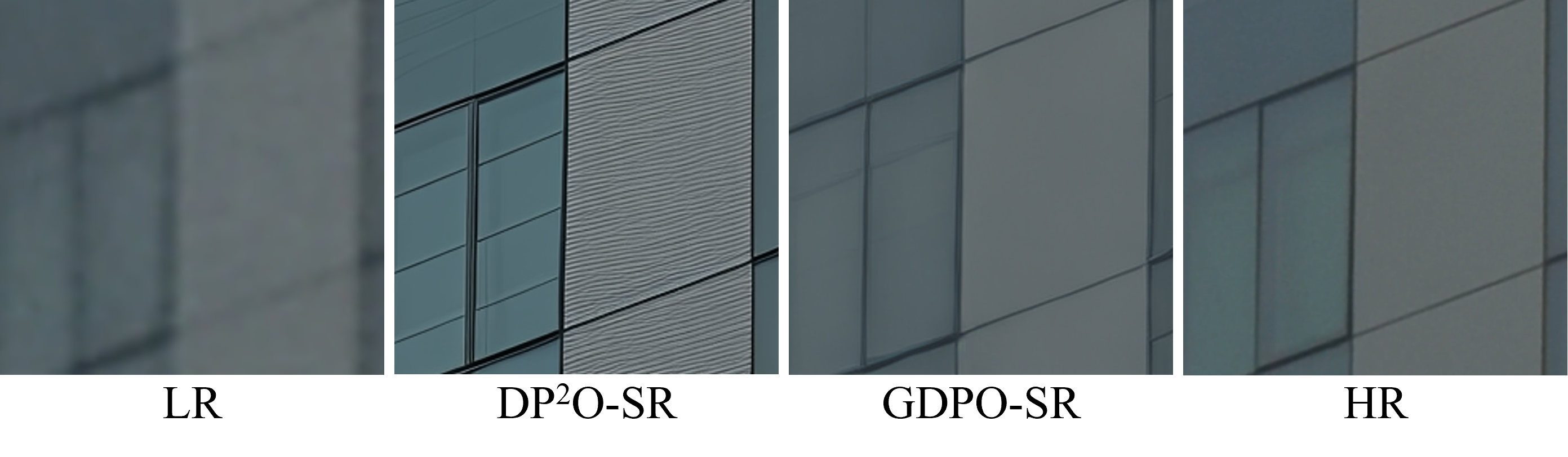}
    \vspace{-4mm}
    \caption{Visual comparison with DP$^2$O-SR.}
    \vspace{-3mm}
    \label{fig:d2p}
\end{figure}

\noindent\textbf{Compared with RL-based methods.} We further compare GDPO-SR with DP$^2$OSR \cite{wu2025dp2osr}, a recent RL-based multi-step Real-ISR method. The quantitative results are presented in Table \ref{tab:rlresults}. We see that DP$^2$O-SR achieves better NR metrics but performs poorly on FR metrics. This suggests that DP$^2$O-SR promotes strong generative capability in the price of compromising image fidelity. In contrast, our GDPO-SR, as a one-step diffusion model, achieves a more balanced performance, improving fidelity while maintaining competitive perceptual quality. The qualitative comparison in Fig.~\ref{fig:d2p} confirms this observation. DP$^2$O-SR generates richer but inconsistent details with the HR image. Our GDPO-SR produces more faithful results.

\noindent\textbf{Complexity Analysis.} Table \ref{tab:flops} provides a comparison of parameter counts (Para.), running time, and FLOPs among different SD-based methods. The running time is the average time over processing 100 $512\times512$ images. From the results, it can be seen that GDPO-SR has the same inference time and FLOPs as OSEDiff, indicating that the introduction of noise and the application of GDPO do not add any computational overhead. Compared with InvSR, although GDPO-SR has slightly more parameters (1.77B vs. 1.33B), it achieves higher inference efficiency (0.11s vs. 0.12s) and smaller FLOPs (2.27T vs. 2.40T).

\subsection{Ablation Study}

\begin{table}
	\centering
        \caption{Ablation studies on GDPO on the RealSR dataset.}
        \vspace{-3mm}
	\resizebox{1.0\linewidth}{!}{
		\begin{tabular}{c|cccccc}
			\toprule
			Method & PSNR↑ & FID↓ & DISTS↓ & MUSIQ↑ & AFINE↓ \\ \midrule
       NAOSD & 25.25 & 114.91 & 0.2001 & 69.06 & 20.52 \\
       Diffusion-DPO & 25.41 & 112.87 & 0.2010 & 69.16 & 20.22 \\
       DanceGRPO & 25.10 & 113.74 & 0.2049 & 69.95 & 16.52 \\
\rowcolor{lightpink}  GDPO (ours) & 25.48 & 112.13 & 0.1980 &  69.42 & 18.72\\
	\bottomrule
	\end{tabular}}
        \vspace{-0.5mm}
 \label{tab:ablationgdpo}
 \vspace{-4mm}
\end{table}

\noindent\textbf{Effectiveness of GDPO.} To demonstrate the effectiveness of the proposed GDPO strategy, we conduct comparisons with other reinforcement learning methods, including diffusion-DPO \cite{wallace2024diffusion} and DanceGRPO \cite{xue2025dancegrpo}. The results are shown in Table \ref{tab:ablationgdpo}. Compared with Diffusion-DPO, our GDPO achieves better performance on both FR and NR metrics, highlighting its advantage in leveraging inter-group online samples for balanced optimization. Compared with NAOSD, DanceGRPO enhances NR metrics but leads to a degradation in FR metrics. It indicates that the global likelihood estimation in DanceGRPO cannot capture local image distributions, causing a deterioration in image fidelity. Overall, our GDPO achieves a favorable balance between perceptual quality and fidelity.

\begin{table}
	\centering
        \caption{Ablation studies on ARF on the RealSR dataset.}
        \vspace{-3mm}
	\resizebox{0.9\linewidth}{!}{
		\begin{tabular}{c|ccccc}
			\toprule
			Method & LPIPS↓ & DISTS↓ & MUSIQ↑ & CLIPIQA↑ \\ \midrule
       NAOSD & 0.2689 & 0.2001 & 69.06 & 0.6617 \\
       ARF w/ FR & 0.2642 & 0.1978 & 67.68 & 0.6359 \\
       ARF w/ NR & 0.2866 & 0.2102 & 69.80 & 0.6914 \\
       ARF w/o AW & 0.2660 & 0.1979 & 68.42 & 0.6331 \\
\rowcolor{lightpink}  ARF (ours) & 0.2675 & 0.1980 & 69.42 & 0.6760 \\
	\bottomrule
	\end{tabular}}
 \label{tab:ablationARF}
 \vspace{-3mm}
\end{table}

\noindent\textbf{Effectiveness of ARF.} To verify the effectiveness of different components within ARF, we conducted five ablation experiments in Table \ref{tab:ablationARF}. ARF w/ FR uses only the FR metric in the reward function, while ARF w/ NR uses only the NR metric. ARF w/o AW removes the adaptive weighting by fixing both $\rho_s$ and $\rho_d$ in \cref{eq:reward} to 0.5. Compared with NAOSD, incorporating only the FR metric (“ARF w/ FR”) improves FR metrics such as LPIPS and DISTS but leads to a noticeable drop in NR metrics. In contrast, using only the NR metric (“ARF w/ NR”) enhances perceptual quality while degrading fidelity. These results suggest that relying solely on either FR or NR metrics is insufficient to evaluate the overall quality of the reconstructed images. Besides, compared with ARF, removing the adaptive weighting (“ARF w/o AR”) results in suboptimal performance across all metrics, indicating that a uniform weighting scheme cannot capture spatially varying characteristics of image content. This demonstrates that adaptively combining FR and NR metrics according to image content provides a reliable evaluation of image quality.

\noindent\noindent\textbf{The effect of group size.} We conducted four experiments to investigate the effect of group size ($G$), with G set to 4, 6, and 8, respectively. The results are presented in Table~\ref{tab:ablationG}. It can be observed that as G increases, the model’s generative capability improves, as indicated by the higher NR metric. This suggests that a larger G leads to more diverse sample generation, thereby enhancing the model’s overall generative performance. In contrast, when G is too small, the limited diversity of samples restricts effective preference learning, resulting in minimal performance gains.

\begin{table}
	\centering
        \caption{Ablation studies on Group Size on the RealSR dataset.}
        \vspace{-3mm}
	\resizebox{0.9\linewidth}{!}{
		\begin{tabular}{c|cccc}
			\toprule
			G & PSNR$\uparrow$ & DISTS$\downarrow$ & MANIQA$\uparrow$ & CLIPIQA$\uparrow$ \\
\midrule
NAOSD       & 25.25 & 0.2001 & 0.6459 & 0.6617 \\
GDPO-SR~(G=4) & 25.45 & 0.1996 & 0.6491 & 0.6610 \\
\rowcolor{lightpink} GDPO-SR~(G=6) & 25.48 & 0.1980 & 0.6615 & 0.6760 \\
GDPO-SR~(G=8) & 25.34 & 0.1955 & 0.6619 & 0.6788 \\
	\bottomrule
	\end{tabular}}
 \label{tab:ablationG}
 \vspace{-3mm}
\end{table}

\begin{figure}[t]
    \centering
    \includegraphics[width=0.9\linewidth]{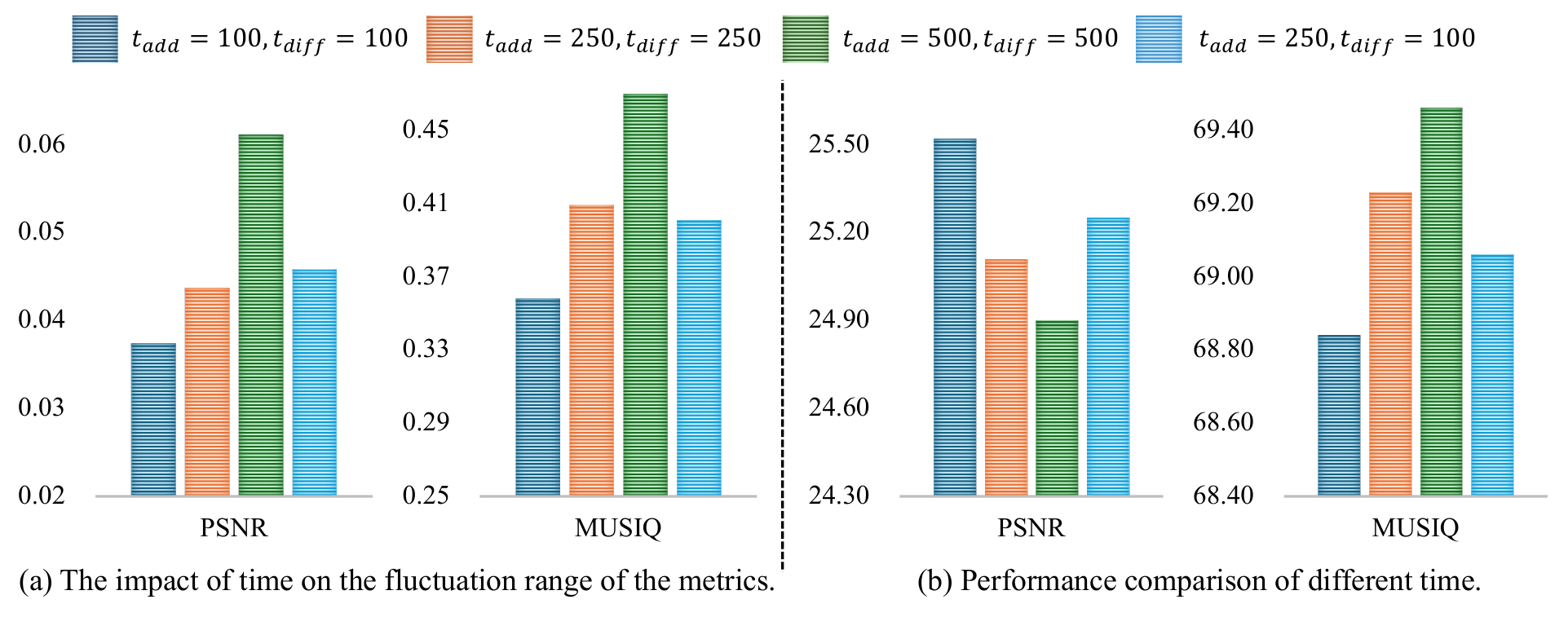}
    \vspace{-4mm}
    \caption{Ablation studies on the timestep setting in NAOSD.}
    \vspace{-5mm}
    \label{fig:ablationtime}
\end{figure}

\noindent\textbf{The timestep in NAOSD.}  To investigate the impact of $t$=($t_{add}$, $t_{diff}$) on NAOSD, we conduct four ablation experiments by settings $t$ as (100,100), (250,250), (500,500) and (250,100). Fig.~\ref{fig:ablationtime}(a) shows the fluctuation range, which is defined as the difference between the maximum and minimum values for each metric, under the four settings. For each setting, we randomly sample 50 outputs per input on the RealSR dataset, yielding 50 values for each metric. As can be seen, when the timestep increases from (100,100) to (500,500), both PSNR and MUSIQ exhibit larger fluctuation ranges. This suggests that a larger $t$ introduces more noise into the generation process, which increases randomness and enhances sample diversity.
Fig. \ref{fig:ablationtime}(b) presents the performance comparison across different $t$. As observed, increasing $t$ enhances the generative ability but compromises fidelity. To address this trade-off, we adopt an unequal-time strategy ($t_{add}=250$, $t_{diff}=100$) to balance fidelity and generative capacity while maintaining a reasonable fluctuation range.

\section{Conclusion}
In this paper, we proposed Group Direct Preference Optimization (GDPO), a novel RL framework for one-step generative image super-resolution. Specifically, we first presented the NAOSD base model with an unequal-timestep strategy, injecting controllable noise into the latent space to produce diverse outputs. Based on NAOSD, we proposed GDPO, which integrated the strengths of DPO and GRPO to perform online preference optimization using multiple generated samples. To effectively distinguish quality differences among these samples, we designed an attribute-aware reward function that dynamically balanced fidelity-related and perception-related metrics according to the content of smooth and textured regions. Extensive experiments demonstrated that GDPO-SR effectively enhanced both visual quality and structural fidelity, achieving superior performance over existing Real-ISR methods.

\noindent\textbf{Limitations}. While GDPO achieves better performance than DPO, it needs to generate multiple outputs for each input during training, which increases training overhead. In addition, ARF provides more reasonable reward scores by adaptively combining FR and NR metrics according to image content, but it remains a manually designed heuristic reward. How to design a reward aligning better with human visual perception deserves more investigation.

% \newpage
{
    \small
    \bibliographystyle{ieeenat_fullname}
    \bibliography{main}
}

\newpage
\onecolumn

% \onecolumn[
%     \centering
%     \Large
%     \textbf{\thetitle}\\
%     \vspace{0.5em}Supplementary Material \\
%     \vspace{1.0em}
% ]

\customtitle{\Large{ Supplementary Material to ``GDPO-SR: Group Direct Preference Optimization \\
\vspace{0.5em} for One-Step Generative Image Super-Resolution''}}

\appendix
\noindent
The following materials are provided in this supplementary file (unless otherwise specified, GDPO-SR adopts $t_{add}=250$ and $t_{diff}=100$ during inference):

\vspace{+1mm}
\begin{enumerate}[leftmargin=1.5em]
    \item[\ref{sec:suppl:noise}\textcolor{cvprblue}{\textbf{.}}] The average performance comparison between the baseline and GDPO-SR on the RealSR dataset;
    \item[\ref{sec:suppl:control}\textcolor{cvprblue}{\textbf{.}}] Controlling the generative capability of GDPO-SR through the timestep $t_{add}$;
    \item[\ref{sec:suppl:ablation1}\textcolor{cvprblue}{\textbf{.}}] Ablation study on FR reward metrics ;
    \item[\ref{sec:suppl:ablation}\textcolor{cvprblue}{\textbf{.}}] Ablation study on sample generation methods in RL;
    \item[\ref{sec:suppl:results}\textcolor{cvprblue}{\textbf{.}}] More visual comparisons (referring to Sec. 5.2 in the main paper);
    \item[\ref{sec:suppl:gan}\textcolor{cvprblue}{\textbf{.}}] Comparisons with GAN-based methods.
\end{enumerate}

\section{Average performance comparison}
\label{sec:suppl:noise}
The model exhibits varying performance under different sampling noise. To ensure a fair and reliable comparison, we conduct 50 independent experiments on the RealSR dataset and evaluate the average performance of NAOSD and GDPO-SR. The results, presented in Table~\ref{tab:ave}, show that GDPO-SR achieves superior average performance compared to NAOSD, indicating that the overall performance of the model is enhanced after reinforcement learning.

\begin{table*}[!b]
\caption{Performance comparison (averaged over 50 stochastic runs) on the RealSR dataset. Arrows denote if higher (↑) or lower (↓) values represent better performance. The best results are highlighted in {\color[HTML]{FF0000} \textbf{red}}.}
\centering
	\resizebox{\linewidth}{!}{
\begin{tabular}{c|c c c c c c c c c }
\toprule
 Method    & PSNR↑                                 & SSIM↑                                   & LPIPS↓                                  & FID↓                           & DISTS↓                         & MANIQA↑                        & MUSIQ↑                        & CLIPIQA↑                       & AFINE↓                        \\ \hline
\hhline{----------} 
 NAOSD & {\color[HTML]{333333} 25.27}          & {\color[HTML]{FF0000}\textbf{ 0.7354}}          & {\color[HTML]{333333} 0.2685}          & {\color[HTML]{333333} 116.67} & {\color[HTML]{333333} 0.1997} & {\color[HTML]{333333} 0.6458} & {\color[HTML]{333333} 68.85} & {\color[HTML]{333333} 0.6620} & {\color[HTML]{333333} 19.28} \\
 GDPO-SR & {\color[HTML]{FF0000}\textbf{ 25.49}}          & {\color[HTML]{333333} 0.7336}          & {\color[HTML]{FF0000}\textbf{ 0.2673} }         & {\color[HTML]{FF0000}\textbf{ 116.03}} & {\color[HTML]{FF0000}\textbf{ 0.1978}} & {\color[HTML]{FF0000}\textbf{ 0.6607}} & {\color[HTML]{FF0000}\textbf{ 69.21}} & {\color[HTML]{FF0000}\textbf{ 0.6766}} & {\color[HTML]{FF0000}\textbf{ 18.19}} \\
\hline
\hhline{----------}  
\end{tabular}}
\label{tab:ave}
\vspace{-2mm}
\end{table*}

\section{Control of Generative Capability}
\label{sec:suppl:control}
The generative capability of GDPO-SR can be controlled by adjusting the diffusion timestep $t_{add}$ during inference. This adjustment allows the model to balance fidelity and realism according to different requirements. The quantitative results on the RealSR dataset are presented in Table \ref{suptab:time}. Note that the GDPO-SR model is fixed, only $t_{add}$ is adjusted during inference, and all variants are fed with the identical noise for the same input. As can be seen, $t_{add}$ provides an effective way to control the model’s generative capability: larger $t_{add}$ lead to higher no-reference metric scores, indicating stronger generative capability. 

\section{Ablation Study on FR reward metrics } 
\label{sec:suppl:ablation1}
We conducted four experiments on the RealSR dataset to investigate the effect of different FR metrics as reward function, including PSNR, LPIPS, and their combination PSNR+LPIPS. As shown in Table~\ref{tab:ablationFR}, using these FR metrics as rewards consistently improves performance. Since LPIPS is a semantic perceptual metric that is less sensitive to pixel-wise errors, using it alone as the reward results in slight PSNR gains. The combination PSNR+LPIPS alleviates this limitation, achieving a better balance.

\begin{table}[!b]
	\centering
        \caption{Ablation studies on FR metrics on the RealSR dataset.}
        \vspace{-3mm}
	\resizebox{0.5\linewidth}{!}{
		\begin{tabular}{c|cccc}
			\toprule
			FR & PSNR$\uparrow$ &  DISTS$\downarrow$ & MANIQA$\uparrow$ & CLIPIQA$\uparrow$ \\
\midrule
NAOSD       & 25.25 & 0.2001 & 0.6459 & 0.6617 \\
\rowcolor{lightpink} PSNR        & 25.48 & 0.1980 & 0.6615 & 0.6760 \\
LPIPS       & 25.28 & 0.1997 & 0.6605 & 0.6835 \\
PSNR+LPIPS  & 25.33 & 0.1970 & 0.6548 & 0.6701 \\
	\bottomrule
	\end{tabular}}
 \label{tab:ablationFR}
 \vspace{-3mm}
\end{table}

\section{Ablation Study on Sample Generation Methods in RL}
\label{sec:suppl:ablation}

There are various ways to generate multiple samples, such as adjusting the classifier-free guidance (CFG) scale or modifying the diffusion timestep $t_{add}$. In this paper, GDPO-SR generates multiple samples by altering the injected noise, enabling diverse outcomes from a single input. To investigate the effects of different sample generation methods, we conduct an ablation study on the Real-ISR dataset, as shown in Table \ref{suptab:sample}. GDPO-SR-CFG denotes the variant that generates samples by changing the CFG, while GDPO-SR-$t_{add}$ represents the variant by varying $t_{add}$.  As shown in Table \ref{suptab:sample}, both the CFG-based and timestep-based sampling methods lead to notable improvements in no-reference metrics; however, they tend to degrade full-reference metrics. In contrast, generating multiple samples with different noises yields more consistent improvements across both no-reference and full-reference metrics.

This is mainly because, when generating multiple samples, changing the CFG or $t_{add}$ introduces an inherent trade-off between fidelity and perceptual quality rather than improving both simultaneously. As demonstrated in Table \ref{suptab:time}, increasing $t_{add}$ enhances perceptual quality (as reflected by higher no-reference metric scores) but degrades fidelity (as indicated by lower full-reference metric scores), and vise versa. Empirically, adjusting the CFG exhibits a similar trend: higher CFG values tend to improve perceptual quality while reducing fidelity. In contrast, altering the noise can generate richer and more natural variations among samples, including rare but valuable cases that simultaneously achieve high fidelity and high perceptual quality. Due to the existence of these rare samples, generating multiple samples with different noises can simultaneously enhance the model’s generative capability and fidelity.

\begin{table*}[!tb]
\caption{The impact of $t_{add}$ on the generation capability of GDPO-SR. Arrows denote if higher (↑) or lower (↓) values represent better performance. The best and second best results are highlighted in {\color[HTML]{FF0000} \textbf{red}} and {\color[HTML]{6434FC} \textbf{blue}}, respectively.}
\centering
	\resizebox{1\linewidth}{!}{
\begin{tabular}{c|cccccccc }
\toprule
Method    & PSNR↑                                 & SSIM↑                                   & LPIPS↓                                                             & DISTS↓                         & MANIQA↑                        & MUSIQ↑                        & CLIPIQA↑                       & AFINE↓                        \\ \hline
GDPO-SR ($t_{add}$=150)   & { \color[HTML]{FF0000}\textbf{25.85}}          & {\color[HTML]{FF0000} \textbf{0.7499}}          & {\color[HTML]{FF0000} \textbf{0.2541}}        & {\color[HTML]{FF0000} \textbf{0.1938}} & { {0.6521}} & { {68.09}} & { {0.6331}} & { {19.89}}\\
GDPO-SR ($t_{add}$=250)      & { \color[HTML]{6434FC}\textbf{25.48}}          & { \color[HTML]{6434FC}\textbf{0.7328}}          & {\color[HTML]{6434FC} \textbf{0.2675}}    & { \color[HTML]{6434FC}\textbf{0.1980}} & {\color[HTML]{6434FC} \textbf{0.6615}} & {\color[HTML]{6434FC} \textbf{69.42}} & { \color[HTML]{6434FC}\textbf{0.6760}} & {\color[HTML]{6434FC} \textbf{17.73}}\\ 
GDPO-SR ($t_{add}$=350)       & { {24.84}}          & { {0.7074}}          & { {0.2900}}          & { {0.2088}} & {\color[HTML]{FF0000} \textbf{0.6646}} & { \color[HTML]{FF0000}\textbf{70.14}} & { \color[HTML]{FF0000}\textbf{0.6994}} & {\color[HTML]{FF0000} \textbf{17.41}} \\  \bottomrule
\end{tabular}}
\label{suptab:time}
\end{table*}

\begin{table*}[!tb]
\caption{Ablation study on sample generation methods on the Real-ISR dataset. Arrows denote if higher (↑) or lower (↓) values represent better performance.}
\centering
	\resizebox{1\linewidth}{!}{
\begin{tabular}{c|cccccccc }
\toprule
Method    & PSNR↑                                 & SSIM↑                                   & LPIPS↓                                                             & DISTS↓                         & MANIQA↑                        & MUSIQ↑                        & CLIPIQA↑                       & AFINE↓                        \\ \hline
NAOSD &25.25          & 0.7346         & 0.2689           & 0.2001 & 0.6459 & 69.06 &  0.6617 & 18.90 \\ 
GDPO-SR (CFG)   & 25.06 & 0.7174  & 0.2782  & 0.2049 & 0.6543 & 69.47 & 0.6815 & 19.61 \\
GDPO-SR ($t_{add}$)      & 25.29    & 0.7294          & 0.2871           & 0.2133 & 0.6477 & 69.46 & 0.6766 & 17.12                     \\ 
\rowcolor{lightpink}  GDPO-SR      & 25.48    & 0.7328          & 0.2675           & 0.1980 & 0.6615 & 69.42 & 0.6760 & 17.73 \\  \bottomrule
\end{tabular}}
\label{suptab:sample}
\end{table*}

\begin{table*}[!]
           \centering
 \caption{
        Quantitative comparison  between GDPO-SR and  the state-of-the-art GAN-based Real-ISR methods on synthetic and  real-world datasets. The best and second best results are highlighted in {\color[HTML]{FF0000} \textbf{red}} and {\color[HTML]{6434FC} \textbf{blue}}, respectively. Arrows denote if higher (↑) or lower (↓) values represent better performance.}
 
   \resizebox{1.0\linewidth}{!}{
\begin{tabular}{c|c|ccccccccc}
\hline
Datasets                   & Methods                                                  & PSNR↑&SSIM↑& LPIPS↓& FID↓& DISTS↓                                    & MANIQA↑ & MUSIQ↑                                  & CLIPIQA↑  & AFINE ↓                                                \\

\hline        
& RealESRGAN                                            & {\color[HTML]{6434FC} \textbf{24.29}}&{\color[HTML]{FF0000} \textbf{0.6371}}&{\color[HTML]{6434FC} \textbf{0.3112}}& {\color[HTML]{6434FC} \textbf{37.64}}&{\color[HTML]{6434FC} \textbf{0.2141}} &{\color[HTML]{6434FC} \textbf{0.5501}}& 61.06& {\color[HTML]{6434FC} \textbf{0.5277}}&    56.20     \\
                           
& BSRGAN                                         & {\color[HTML]{FF0000} \textbf{24.58}}& 0.6269&0.3351& 44.23 & 0.2275& 0.5247&{\color[HTML]{6434FC} \textbf{61.20}} & 0.5071&   62.21      \\
                           
& LDL                                              & 23.83& {\color[HTML]{6434FC} \textbf{0.6344}}&0.3256& 42.29&0.2227 & 0.5350& 60.04& 0.5180&   {\color[HTML]{6434FC} \textbf{53.19}}      \\

\rowcolor{lightpink}
 \multirow{-4}{*}{DIV2K-val}& GDPO-SR      & 23.92                                & 0.6117          & {\color[HTML]{FF0000} \textbf{0.2897}}          & {\color[HTML]{FF0000} \textbf{26.44}}  & {\color[HTML]{FF0000} \textbf{0.1965}} & {\color[HTML]{FF0000} \textbf{0.6423}} & {\color[HTML]{FF0000} \textbf{68.80}} & {\color[HTML]{FF0000} \textbf{0.6929}} & {\color[HTML]{FF0000} \textbf{41.56}} \\
\hline          

&RealESRGAN                                     & {\color[HTML]{6434FC} \textbf{25.69}} & {\color[HTML]{6434FC} \textbf{0.7616}} & 0.2727 & {\color[HTML]{6434FC} \textbf{135.18}} & {\color[HTML]{6434FC} \textbf{0.2063}} & {\color[HTML]{6434FC} \textbf{0.5487}} & 60.18     &    0.4449                            &    39.53                    \\
& BSRGAN                                               & {\color[HTML]{FF0000} \textbf{26.39}}& {\color[HTML]{FF0000} \textbf{0.7654}} & {\color[HTML]{FF0000} \textbf{0.2670}} & 141.28 & 0.2121 & 0.5399  &   {\color[HTML]{6434FC} \textbf{63.21}}     &   {\color[HTML]{6434FC} \textbf{0.5001}}            &       45.69        \\
& LDL                                              & 25.28 & 0.7567 & 0.2766 &  142.71 & 0.2121 & 0.5485 &  60.82 &  0.4477                               & {\color[HTML]{6434FC} \textbf{38.43}} \\
\rowcolor{lightpink}
\multirow{-4}{*}{RealSR}  & GDPO-SR & 25.48         & 0.7328          & {\color[HTML]{6434FC} \textbf{0.2675}}          & {\color[HTML]{FF0000} \textbf{112.13}} & {\color[HTML]{FF0000} \textbf{0.1980}} & {\color[HTML]{FF0000} \textbf{0.6615}} & {\color[HTML]{FF0000} \textbf{69.42}} & {\color[HTML]{FF0000} \textbf{0.6760}} & {\color[HTML]{FF0000} \textbf{17.73}}\\
  \hline 
  
& RealESRGAN                                               & {\color[HTML]{6434FC} \textbf{28.64}} & {\color[HTML]{6434FC} \textbf{0.8053}} & {\color[HTML]{6434FC} \textbf{0.2847}} & {\color[HTML]{6434FC} \textbf{147.62}} & {\color[HTML]{FF0000} \textbf{0.2089}} & 0.4907   &  54.18      & 0.4422                     & 39.97                 \\
& BSRGAN                                              & {\color[HTML]{FF0000} \textbf{28.75}} & 0.8031 & 0.2883 & 155.63 &  0.2142 & 0.4878     &  {\color[HTML]{6434FC} \textbf{57.14}}      & {\color[HTML]{6434FC} \textbf{0.4915}}                           &    44.87            \\
& LDL                                           & 28.21 & {\color[HTML]{FF0000} \textbf{0.8126}} & {\color[HTML]{FF0000} \textbf{0.2815}} & 155.53 &  0.2132 & {\color[HTML]{6434FC} \textbf{0.4914}} & 53.85   & 0.4310                 & {\color[HTML]{6434FC} \textbf{39.06}}                 \\
\rowcolor{lightpink}
\multirow{-4}{*}{DrealSR}& GDPO-SR & 28.18 & 0.7839 & 0.2851 & {\color[HTML]{FF0000} \textbf{138.87}} & {\color[HTML]{6434FC} \textbf{0.2112}} & {\color[HTML]{FF0000} \textbf{0.6180}} & {\color[HTML]{FF0000} \textbf{65.63}} & {\color[HTML]{FF0000} \textbf{0.7020}} & {\color[HTML]{FF0000} \textbf{18.72}}\\ \hline
  
\end{tabular}
}
\label{tab:comparison_gan}
\end{table*}

\section{More Visual Comparisons}
\label{sec:suppl:results}

\begin{figure*}[t]
    \centering
    \includegraphics[width=1.0\linewidth]{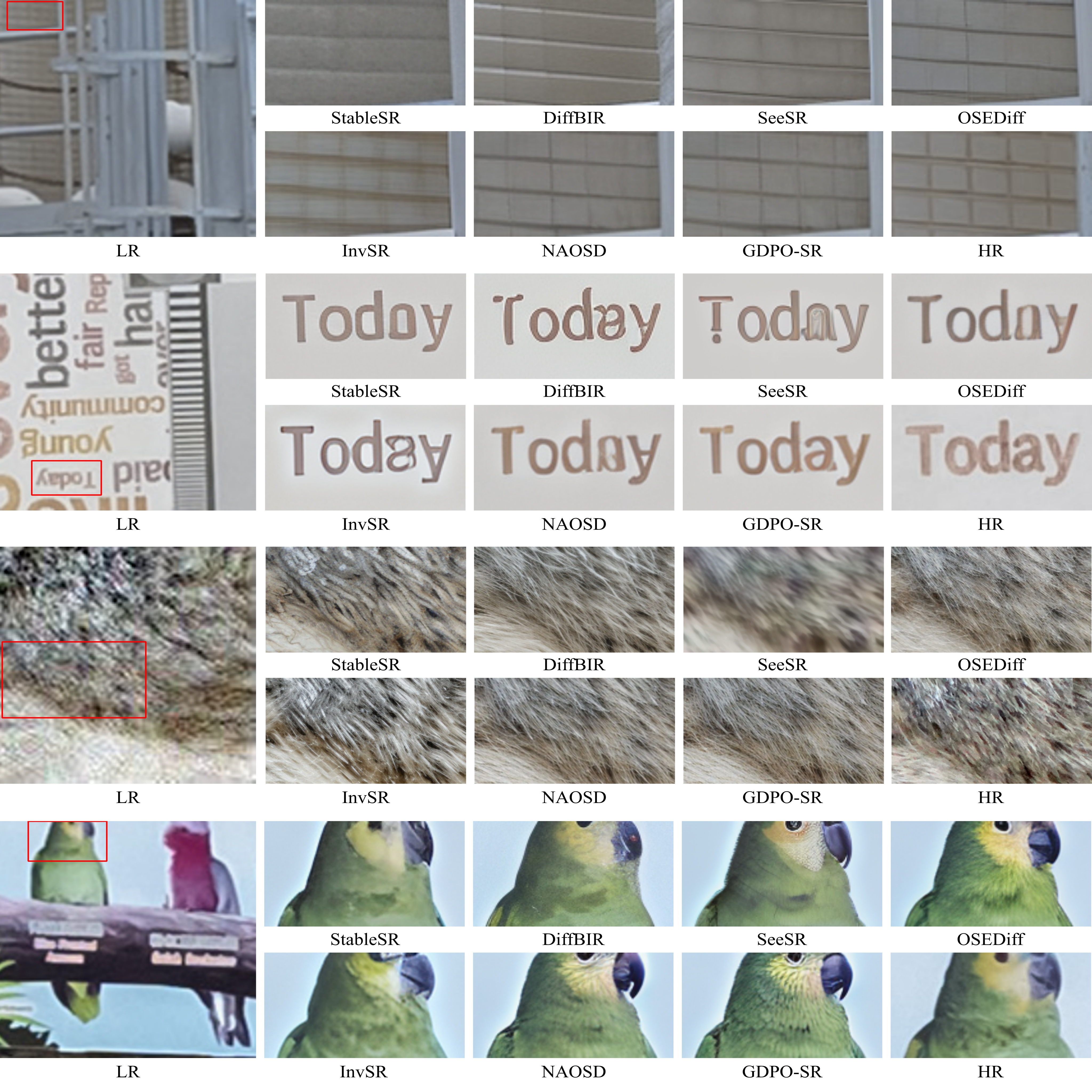}
    \vspace{-4mm}
    \caption{Visual comparison with SD-based Real-ISR methods. Please zoom in for a better view.}
    \label{fig:supvisualresults}
    \vspace{-4mm}
\end{figure*}

We provide more visual comparisons in Fig. \ref{fig:supvisualresults} to demonstrate the effectiveness of GDPO-SR. Firstly, compared with the base model NAOSD, the post-training method GDPO-SR reconstructs sharper and clearer textures (as shown in the first and second cases) and generates finer details (as seen in the third and fourth cases). Secondly, compared to other advanced SD-based methods, GDPO-SR also exhibits notable advantages. For instance, in the second case, GDPO-SR successfully restores the characters “a” and “y”, whereas other methods struggle to produce accurate shapes. Overall, GDPO-SR delivers sharper structures and more natural details, demonstrating strong robustness and generalization in real-world scenarios.

\section{Comparisons with GAN-based Methods}
\label{sec:suppl:gan}

\begin{figure*}[b]
    \vspace{-10mm}
    \centering
    \includegraphics[width=1.0 \linewidth]{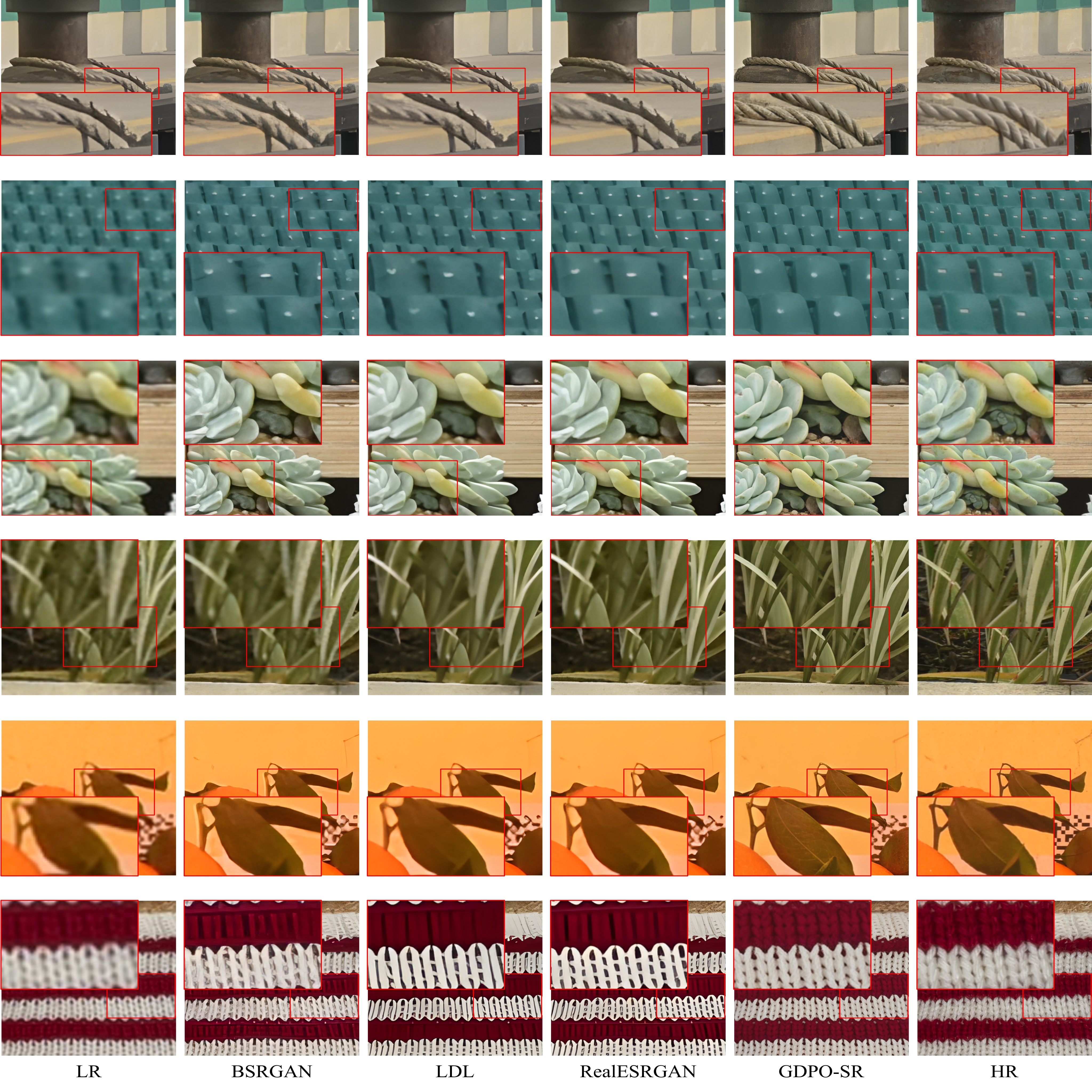}
    \vspace{-4mm}
    \caption{Visual comparison with GAN-based Real-ISR methods. Please zoom in for a better view.}
    \label{fig:supgan}
\end{figure*}

We compare GDPO-SR with three representative GAN-based Real-ISR methods: RealESRGAN \cite{wang2021real}, BSRGAN \cite{zhang2021designing} and LDL \cite{liang2022details}. The quantitative results are summarized in Table \ref{tab:comparison_gan}. As observed, GDPO-SR achieves the best performance on most no-reference metrics (MANIQA \cite{maniqa}, MUSIQ \cite{musiq}, CLIPIQA \cite{clipiqa}, and AFINE \cite{chen2025toward}) across all the three test datasets (DIV2K-val \cite{div2k}, RealSR \cite{realsr}, and DRealSR \cite{drealsr}). 
For full-reference metrics (\eg, LPIPS, DISTS, and FID), GDPO-SR also delivers competitive results. The visual comparisons are illustrated in Fig. \ref{fig:supgan}. It can be clearly found that the proposed GDPO-SR method can generate more realistic details than those GAN-based methods. These results demonstrate that GDPO-SR effectively balances fidelity and perceptual quality, surpassing GAN-based counterparts in overall visual realism and quantitative performance.

\end{document}